\documentclass[11pt,a4paper]{article}
\usepackage[hyperref]{acl2017}
\usepackage{times}
\usepackage{latexsym}
\usepackage{algorithm} 
\usepackage{algorithmic} 
\usepackage{multirow} 
\usepackage{amsmath}
\usepackage{xcolor}
\usepackage{epsfig}

\usepackage{url}

\aclfinalcopy 

\title{A Generic Online Parallel Learning Framework for Large Margin Models}

\author{Shuming Ma \and Xu Sun \\
MOE Key Laboratory of Computational Linguistics, Peking University\\
School of Electronics Engineering and Computer Science, Peking University\\
\{shumingma, xusun\}@pku.edu.cn\\}

\date{}

\begin{document}
\maketitle
\begin{abstract}
To speed up the training process, many existing systems use parallel technology for online learning algorithms.
However, most research mainly focus on stochastic gradient descent (SGD) instead of other algorithms. We propose
a generic online parallel learning framework for large margin models, and also analyze our framework on popular large margin
algorithms, including MIRA and Structured Perceptron. Our framework is lock-free and easy to implement
on existing systems. Experiments show that systems with our framework can gain near linear speed up by increasing running threads, and with no loss in accuracy.
\end{abstract}

\section{Introduction}

Large margin models have been widely used in natural language processing for faster learning rate and smaller
computational cost. However, the algorithms may still suffer from slow training time when training examples
are extremely massive, the weight vector is large, or the inference process is slow. With parallel algorithms, we can
make better use of our multi-core machine and reduce the time cost of training process.

Unluckily, most studies about parallel algorithms mainly focus on SGD. Recht et.al~\shortcite{Rechtetal2011} first proposed a lock-free parallel SGD
algorithm called HOGWILD. It is a simple and effective algorithm which outperforms
non-parallel algorithms by an order of magnitude. Lian et.al~\shortcite{Lianetal2015} provide theoretical analysis of asynchronous parallel SGD
for nonconvex optimization and a more precise description for lock-free implementation on shared memory system.

Zinkevich et.al~\shortcite{Zinkevichetal2010} proposed a parallel algorithm for multi-machine called Parallel Stochastic Gradient Descent (PSGD). PSGD is an effective parallel approach on
distributed machine but Recht et.al~\shortcite{Rechtetal2011} found that it is not as promising as HOGWILD on a single machine.
Zhao and Li~\shortcite{Zhaoetal2016} propose a fast asynchronous parallel SGD approach with convergence guarantee. The method has a much faster convergence rate than HOGWILD.

To the best of our knowledge, there is no related research about asynchronous parallel method for large margin models. McDonald et.al~\shortcite{Mcdonaldetal2010} proposed a distributed structured perceptron algorithm but it needs multi-machine. We first propose a generic online parallel learning framework for large margin models. In our framework, each thread updates the weight vector independently without any extra operation or lock.
Besides, we analyze the performance of structured perceptron and Margin Infused Relaxed Algorithm (MIRA) in our framework. The contribution of our framework can be outlined as follow:
\begin{itemize}
\item Our framework is generic and suitable for most of the large margin algorithms. It is simple and can be easily implemented on existing systems with large margin models.
\item The framework does not use extra memory. Experiments show that the memory cost is no more than single-thread
algorithm. Besides, the parallel framework works on a shared memory system so we have no need to care about the data exchange.
\end{itemize}

\begin{algorithm}[t]
   \caption{The Generic Parallel Framework}\label{algo1}
\begin{algorithmic}[1]

 \REQUIRE training set $S$ with $N$ samples
 \STATE {\textbf{initialize}: weight vector $w=0, v=0$}
 \FOR{$t=1$ to $T$}
 \STATE {Random shuffle training set $S$}
 \STATE {Split training set into $K$ part $\{S_i\}_{i=1}^{K}$}
 \FORALL {threads parallel}
 \STATE {Inference for update term $\Phi_i$}
 \STATE {Update $w_{i+1}$ with $\Phi_i$}
 \STATE {$v = v + w_{i+1}$}
 \STATE {$i = i + 1$}
 \ENDFOR
 \ENDFOR
 \ENSURE the learned weights $\pmb w^* = \pmb v / (N*T)$

\end{algorithmic}
\end{algorithm}

\section{Generic Parallel Learning Framework}

Suppose we have a training dataset with $N$ samples denoted as $\{(x_{i},y_{i})\}_{i=1}^{N}$, where $x_{i}$ is a sequence (usually a sentence
in natural language processing) and $y_{i}$ is a structure on sequence $x_{i}$ (usually a tag list or a tree). The whole dataset
is trained with $T$ passes. We denote the weight vector as $w$ and after $i^{th}$ update the weight vector will be $w^{i}$.

In each pass, the online learning algorithm will shuffle the dataset after which the weight vector is updated with
each sample in the dataset. Generally, gradient-based algorithms like SGD take a lot of time to compute the gradient. Large margin algorithms like perceptron and MIRA spend most of time in decoding.
Popular approach to speed up inference process is to take
approximate inference instead of exact inference \cite{Huangetal2012}. However, we can parallel the inference process of several samples
and update asynchronously to further accelerate the training process~\cite{Sun2016Asynchronous}.

In our parallel framework, we split the dataset into $k$ parts and then assign these split datasets to $k$ threads. Each thread
updates independently with a shared memory system. After that, we average the weight vector by the number of iterations (not the number of threads as some distributed parallel algorithms). We can see that our approach has no more
computation than large margin algorithms, so it is a simple parallel framework. Since the whole framework
runs on a shared memory system, there are mainly two problems about this framework: First, several threads update
at the same time so it may be closer to minibatch algorithm instead of online learning algorithm intuitively. Whether
the parallel framework will affect the convergence rate of online learning algorithm is a problem. Second, when a thread is working,
the weight vector may be overwritten by other threads. Whether it will lead to divergence is another problem need to be analyzed.

For the first problem, Lian et.al~\shortcite{Lianetal2015} proved that the convergence rate will not be affected under the parallel SGD
framework. Our experiments also support that the convergence rate of large margin algorithms is still the
same in our parallel framework. For the second problem, Recht et.al \shortcite{Rechtetal2011} shows that individual SGD steps only modify
a small part of the decision variable so memory overwrites are rare and barely any error will be made into the
computation. We will also explain this problem on large margin models in Section~\ref{sec:larmar}. Experiments show that
our parallel algorithm is so robust that the interference among threads will not affect the convergence.
In our framework we also average the weight vector by the number of iterations. One reason
is that Collins~\shortcite{Collins2002} explains that averaging parameter helps advoid overfitting. Another advantage is that we can ensure
every update will contribute to the final learned weight vector.

\begin{figure*}[tb]
\centering
\begin{tabular}{@{}c@{}@{}c@{}@{}c@{}@{}c@{}}

\epsfig{file=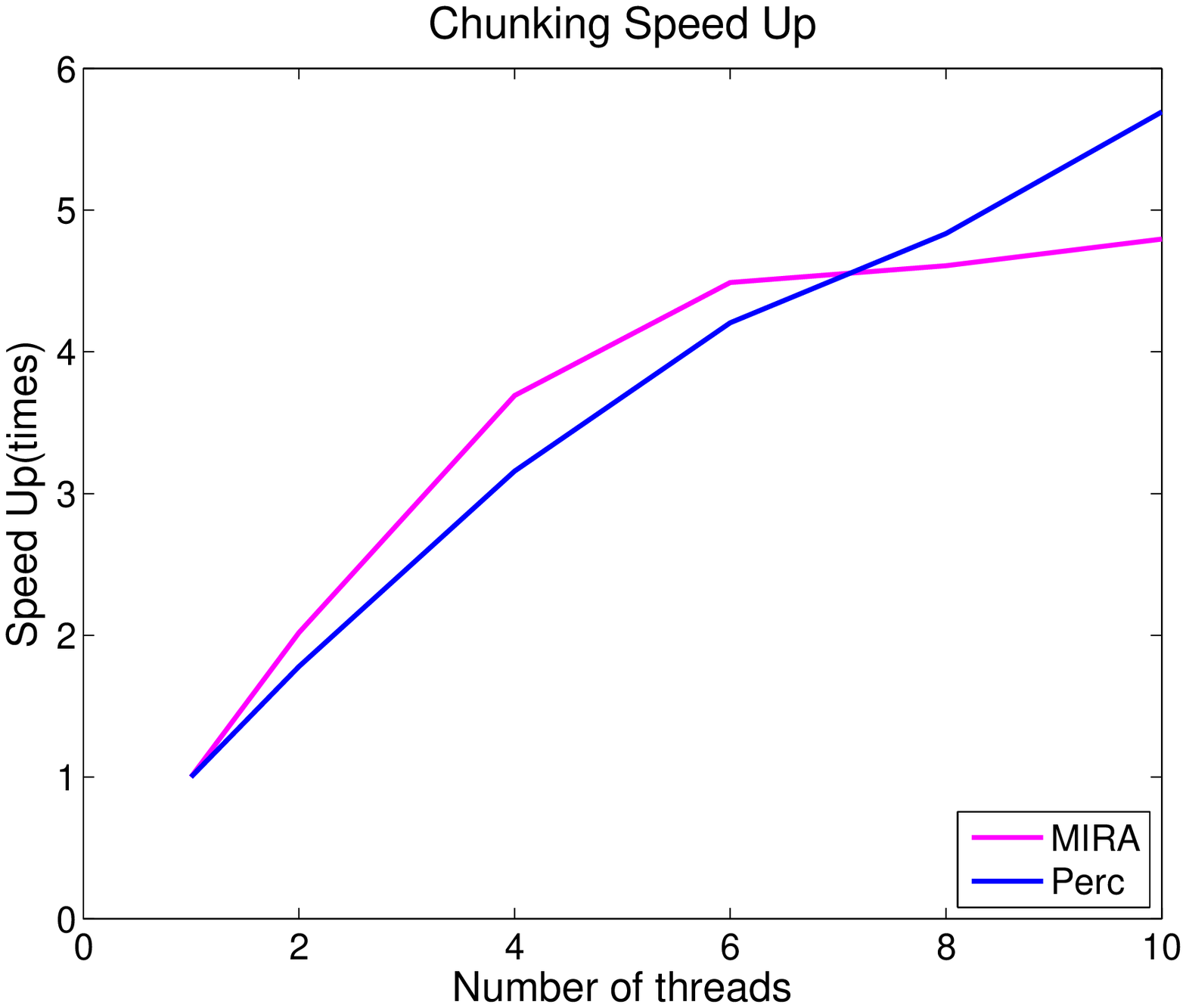,width=0.33\linewidth,clip=} &
\epsfig{file=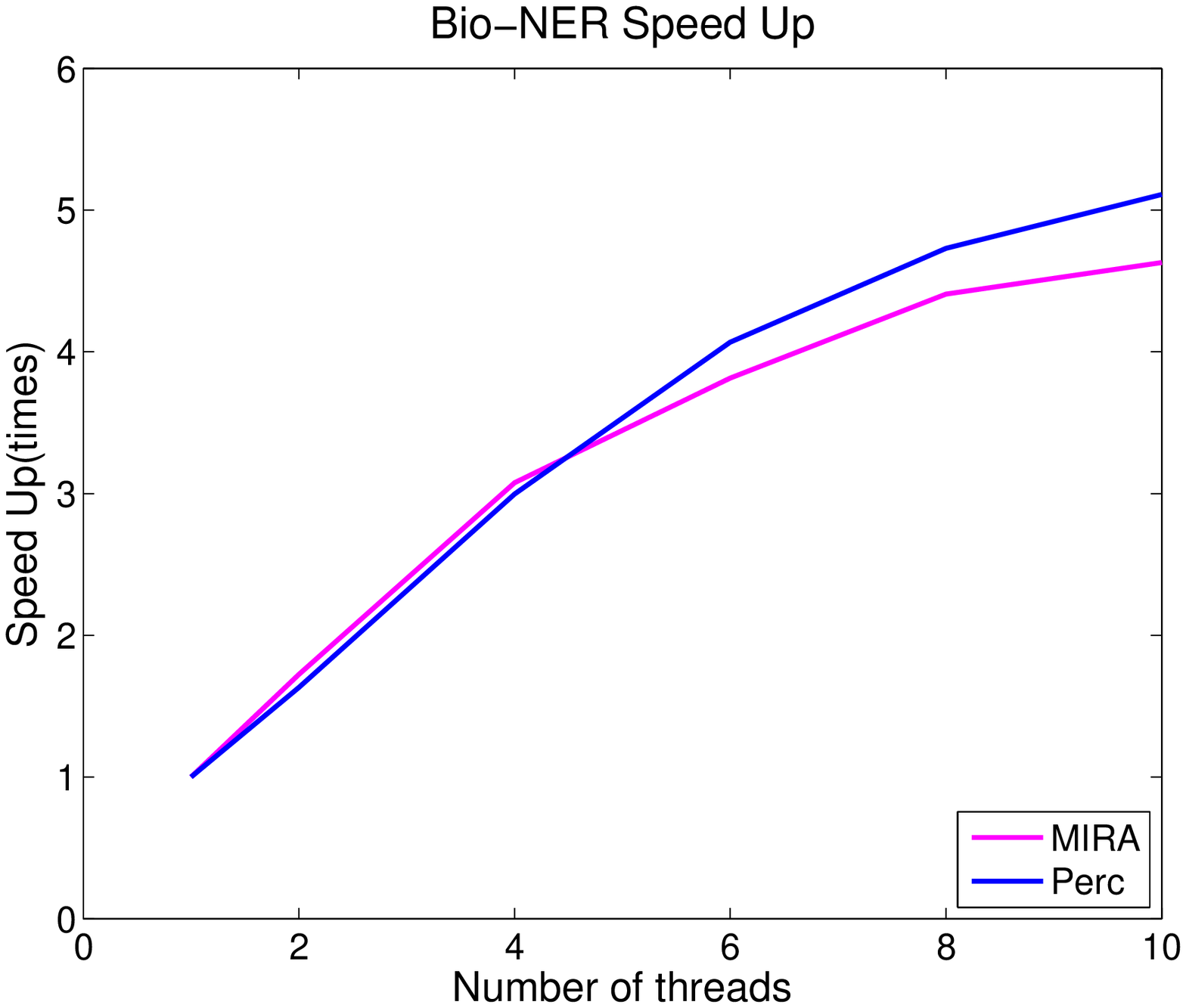,width=0.33\linewidth,clip=} &
\epsfig{file=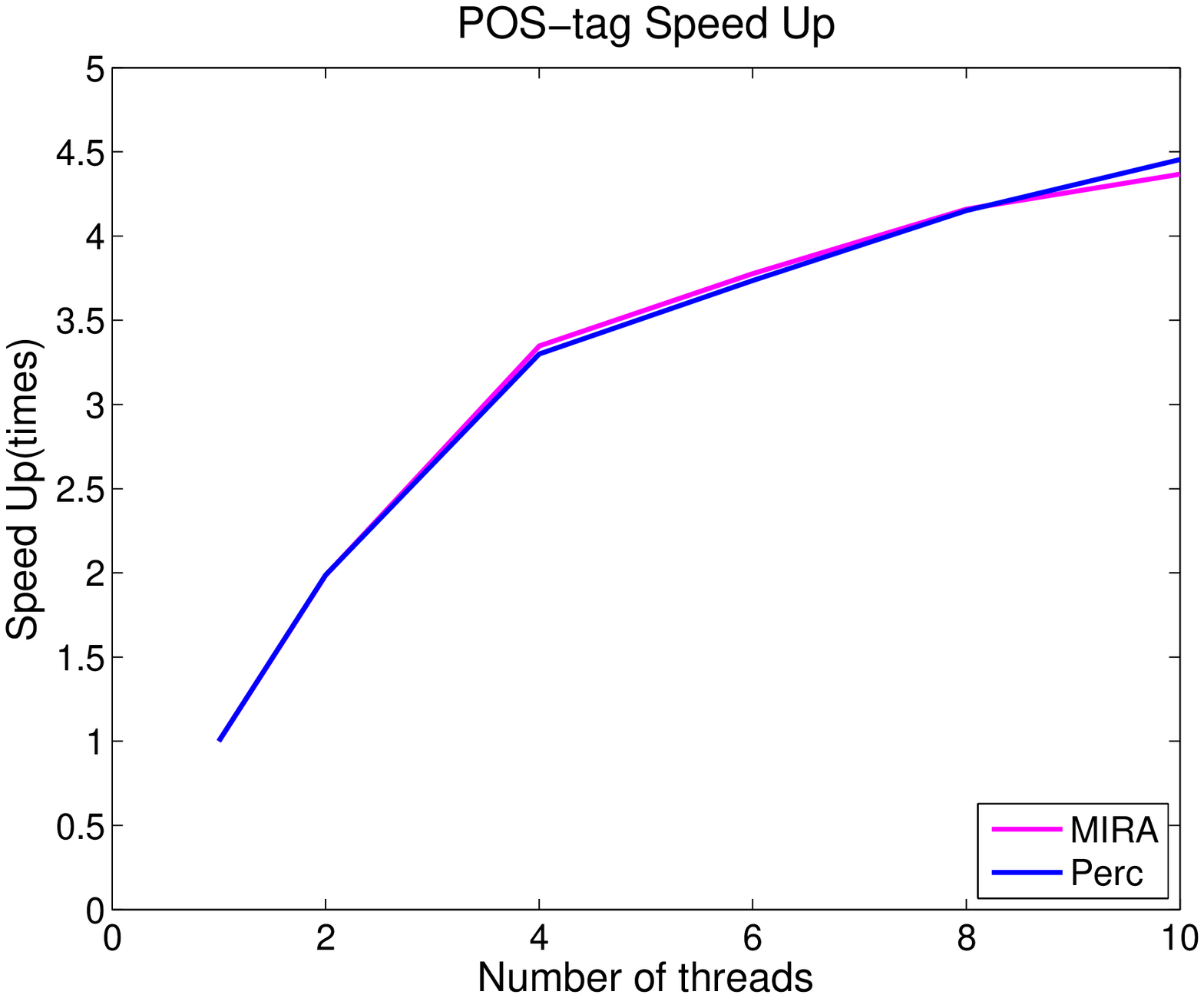,width=0.33\linewidth,clip=} \\

\end{tabular}
\caption{Speed up of our parallel framework.
}\label{fig1}
\vspace{-0.1in}
\end{figure*}

\section{Large Margin Models}\label{sec:larmar}

In this section, we will introduce some popular large margin algorithms and analyze their performance under our parallel framework.

\subsection{MIRA}

Crammer and Singer~\shortcite{Crammer2003} developed a large margin algorithm called MIRA and later extended by Taskar et.al~\shortcite{TaskarEA2004}. The algorithm has been widely used in many popular models \cite{Mcdonaldetal2005}. It tries to minimize $\lVert w \rVert$ so that the margin between output score $s(x,z)$ and correct score $s(x,y)$ is larger than the loss of output structure:
\begin{equation*}
minimize  \lVert w \rVert
\end{equation*}
\begin{equation*}
st. \ \forall z \in GEN(x)\ s(x,y) - s(x,z) \geq L(y,z)
\end{equation*}

During the inference process, the algorithm manages to find out the output structure with the highest score. However, in our parallel framework weight vector can be overwritten so we may not get the 1-best structure. Actually, it does not matter because the binary feature representation is so sparse that the output score is still close to 1-best score. We can say that the margin between output score and correct score is larger than the loss, so the margin between 1-best score and correct score will still satisfy the constrain.

\subsection{Structured Perceptron}

Structured perceptron is first proposed by Collins~\shortcite{Collins2002}. It proves to be an effective and efficient structured prediction algorithm with convergence guarantee for separable data~\cite{SunIJCAI09,tkde/SunML13,SunArXiv2015}. We denote the binary feature representation of sequence $x$ and structure $y$ to be $f(x,y)$. The set of structure candidates for the input sequence $x$ is denoted as $GEN(x)$. Structured perceptron searches the space of $GEN(x)$ and finds the output $z$ with highest score $f(x,z) \cdot w$. The weight vector $w$ then updates with the output $z$.

In our parallel framework, the inference and update of different samples runs parallel. The inference and update for structured perceptron can be descibed as:
\begin{equation}
z= argmax_{t \in GEN(x)}f(x,t) \cdot w
\end{equation}
\begin{equation}
w = w + {f(x,y)} - {f(x,z)}
\end{equation}
During the inference, although the weight vector may be modified, we can still ensure that the score of the output $z$ is higher than that of correct structure $y$. Huang et.al~\shortcite{Huangetal2012} proved that if each update involves a violation (the output has a higher model score than the correct structure), structured perceptron algorithm is bound to converge. Therefore, our parallel framework on structure perceptron is effective theoretically.

\begin{figure*}[tb]
\centering
\begin{tabular}{@{}c@{}@{}c@{}@{}c@{}@{}c@{}}

\epsfig{file=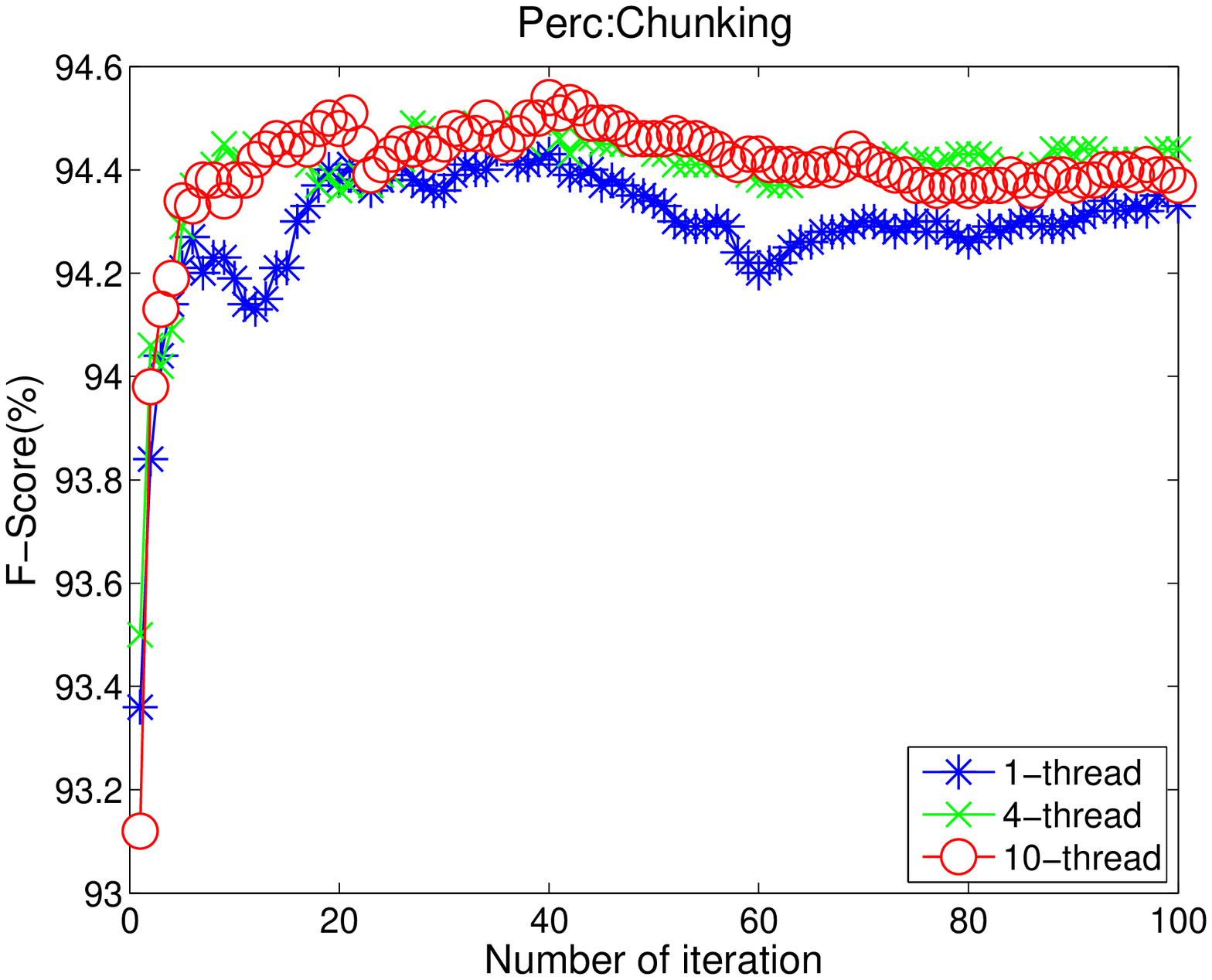,width=0.33\linewidth,clip=} &
\epsfig{file=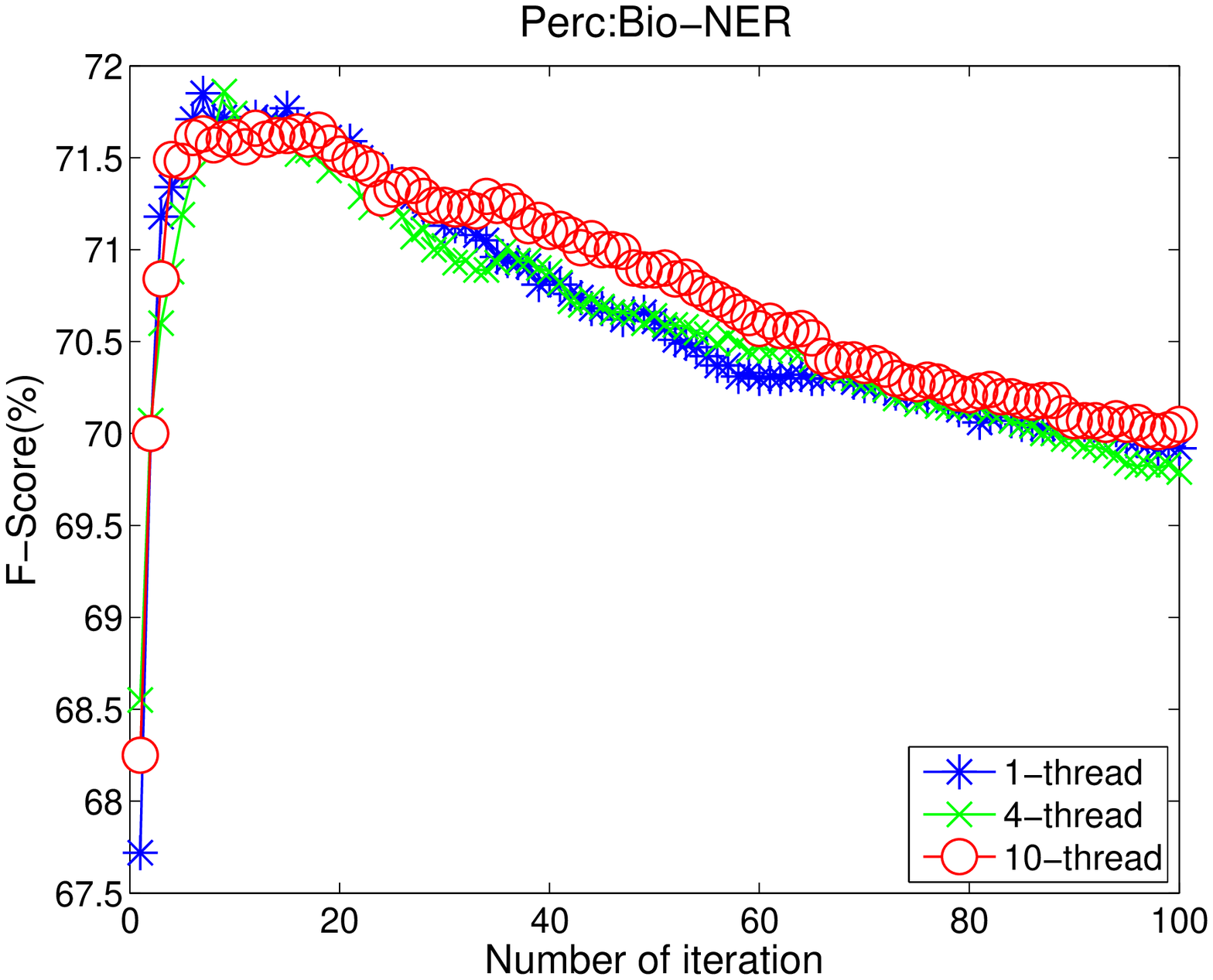,width=0.33\linewidth,clip=} &
\epsfig{file=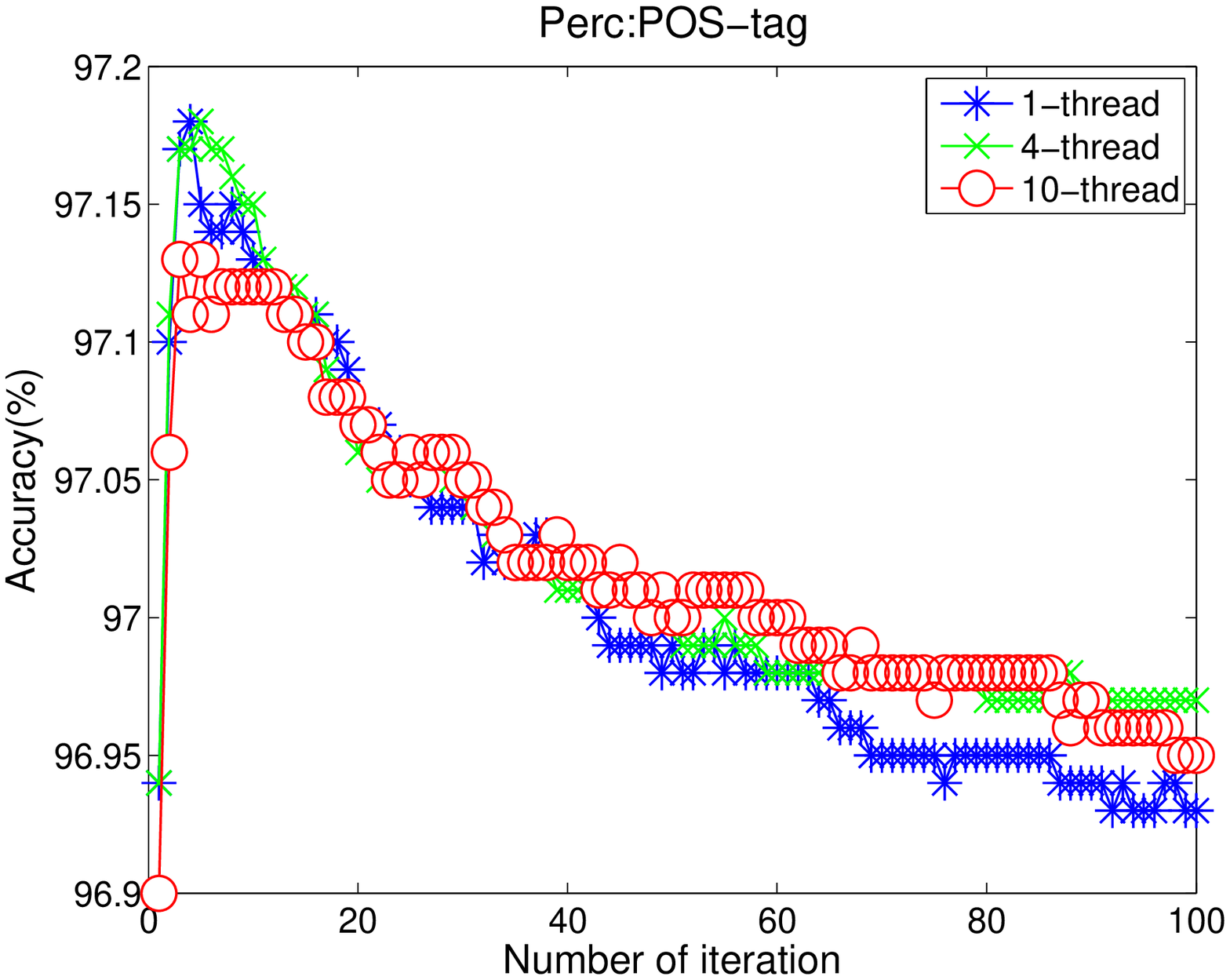,width=0.33\linewidth,clip=} \\

\epsfig{file=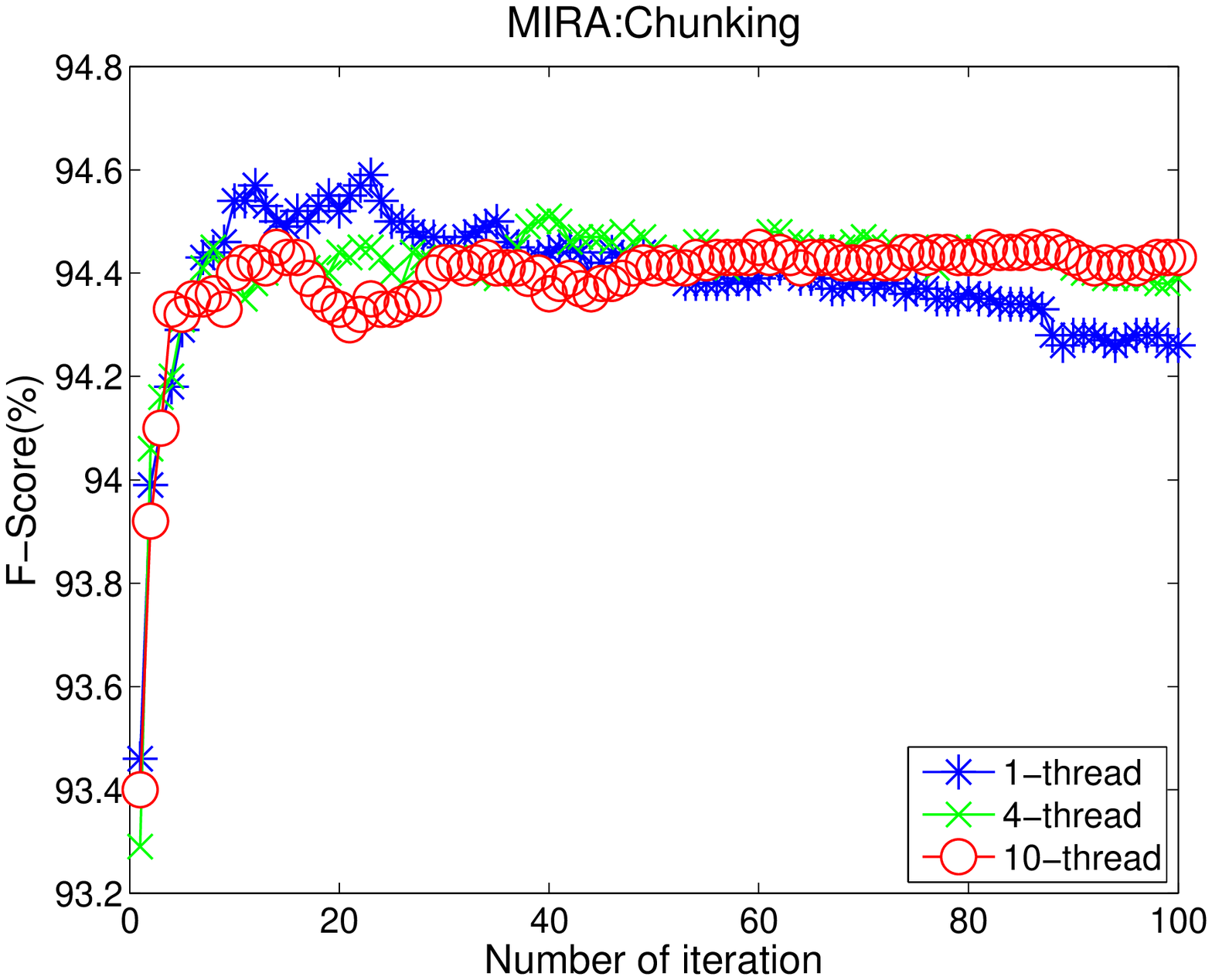,width=0.33\linewidth,clip=} &
\epsfig{file=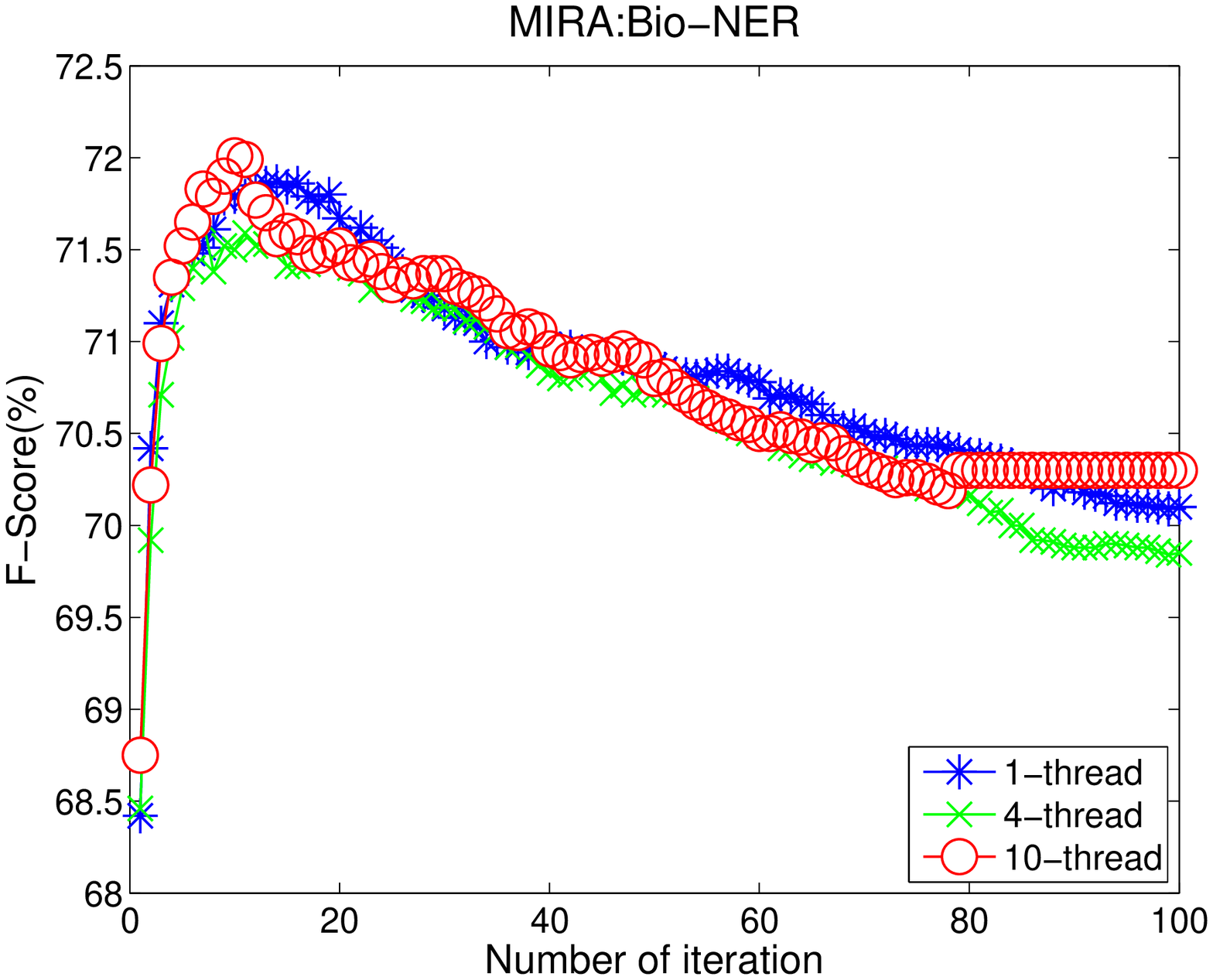,width=0.33\linewidth,clip=} &
\epsfig{file=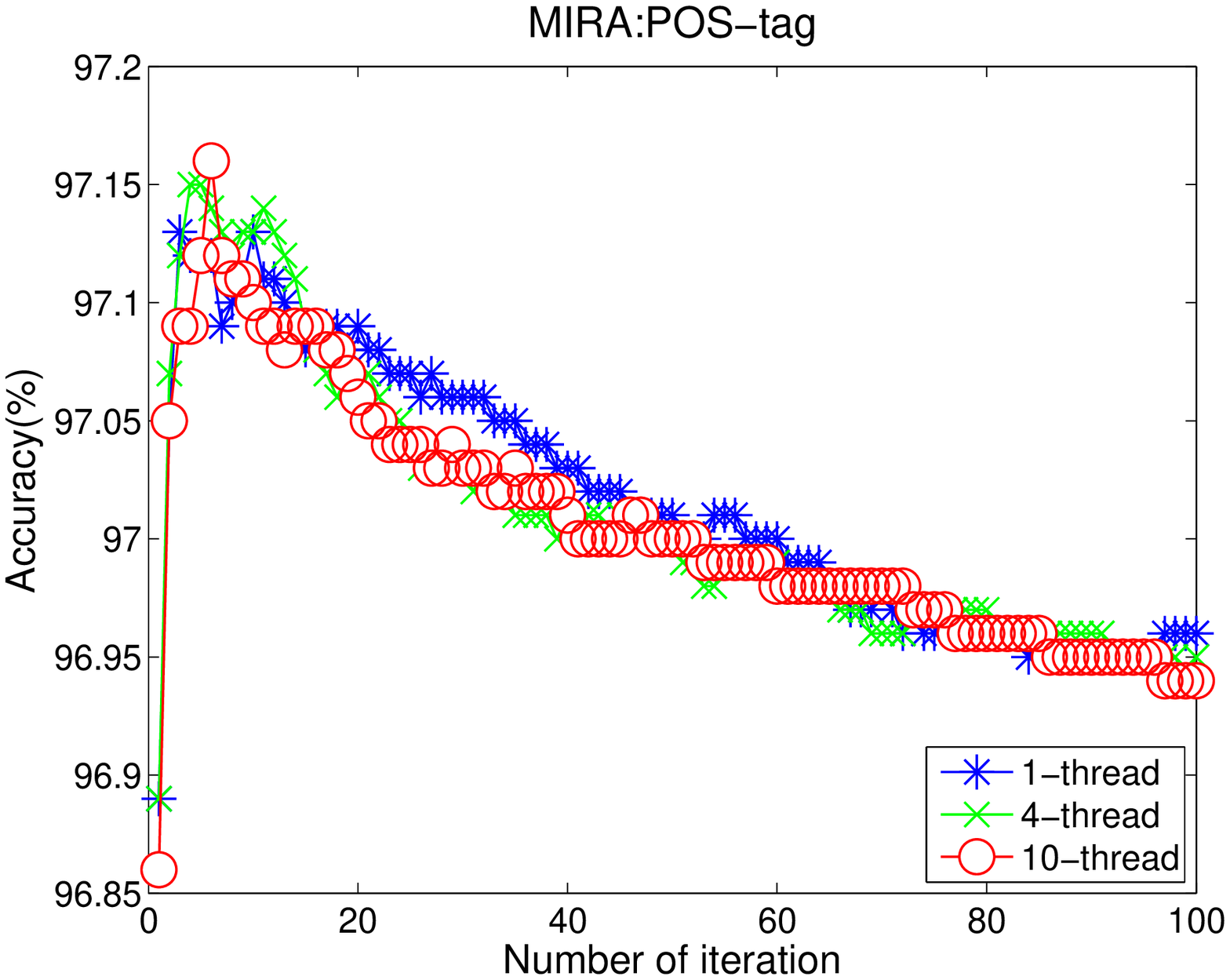,width=0.33\linewidth,clip=} \\

\end{tabular}
\caption{Experiment Results on the benchmark datasets.
}\label{fig2}
\vspace{-0.1in}
\end{figure*}

\section{Experiments}

We compare our parallel framework and non-parallel large margin algorithms on several benchmark datasets.

\subsection{Experiment Tasks}

\textbf{Part-of-Speech Tagging (POS-tag):}
Part of Speech Tagging is a famous and important task in natural language processing. Following the prior work \cite{Collins2002}, we derives the dataset from Penn Wall Street Journal Treebank \cite{Marcusetal1993}. We use sections 0-18 of the treebank as training set while sections 19-21 is development set and sections 22-24 is test set. The selected feature is including unigrams and bigrams of neighboring words as well as lexical patterns of current word \cite{tsuruokaEA2011}. We report the accuracy of output tag as evaluation metric.

\textbf{Phrase Chunking (Chunking):}
In Phrase Chunking task, we tag the words in the sequence to be B, I or O to identify the noun phrases. The dataset is extracted from the CoNLL-2000 shallow-parsing shared task \cite{SangBuchholz2000}. The feature includes word n-grams and part-of-speech n-grams. Our evaluation metric is F-score following prior works.

\textbf{Biomedical Named Entity Recognition (Bio-NER):}
Biomedical Named Entity Recognition is mainly about the recognition of 5 kinds of biomedical named entities. The dataset is from MEDLINE biomedical text corpus. We use word pattern features and part-of speech features in our model \cite{tsuruokaEA2011}. The evaluation metric is F-score.

\subsection{Experiment Setting}

\begin{table}[tb]
\begin{tabular}{|c|c|c|c|c|}
\hline
\multicolumn{2}{|c|}{Number of threads} & 1 & 4 & 10  \\
\hline
\multirow{2}{*}{Chunking} & Perc & 94.40 & 94.46 & \textbf{94.50} \\
\cline{2-5}
& MIRA & \textbf{94.56} & 94.50 & 94.53 \\
\hline
\multirow{2}{*}{Bio-NER} & Perc & 71.83 & \textbf{71.90} & 71.80 \\
\cline{2-5}
& MIRA & 71.75 & 71.70 & \textbf{71.91} \\
\hline
\multirow{2}{*}{POS-tag} & Perc & 97.17 & \textbf{97.18} & 97.10 \\
\cline{2-5}
& MIRA & 97.13 & \textbf{97.15} & \textbf{97.15} \\
\hline
\end{tabular}
\caption{Accuracy/F-score of baseline and our framework.} \label{table1}
\end{table}

\begin{table}[tb]
\begin{tabular}{|c|c|c|c|c|}
\hline
\multicolumn{2}{|c|}{Number of threads} & 1 & 4 & 10  \\
\hline
\multirow{2}{*}{Chunking} & Perc & 1.0x & 3.0x & \textbf{5.5x} \\
\cline{2-5}
& MIRA & 1.0x & 3.7x & \textbf{4.7x} \\
\hline
\multirow{2}{*}{Bio-NER} & Perc & 1.0x & 3.0x & \textbf{5.0x} \\
\cline{2-5}
& MIRA & 1.0x & 3.0x & \textbf{4.6x} \\
\hline
\multirow{2}{*}{POS-tag} & Perc & 1.0x & 3.3x & \textbf{4.4x} \\
\cline{2-5}
& MIRA & 1.0x & 3.4x & \textbf{4.4x} \\
\hline
\end{tabular}
\caption{Speed up of our framework.} \label{table2}
\end{table}

We implement our parallel framework with large margin algorithms, including structured perceptron and MIRA on above benchmark datasets. We use development set to tune the learning rate $\alpha_0$ and L2 regularization. The final learning rate $\alpha_0$ is set as 0.02, 0.05, 0.005 for above three tasks, and L2 regularization is 1, 0.5, 5.

We implement a single-thread framework as our baseline. The setting of baseline is totally the same as our proposed framework. For fair comparison, we also average the parameters of the baseline. We run our parallel framework up to 10 threads following prior work \cite{Rechtetal2011}. We compare our parallel framework with baseline in accuracy/F-score and time cost. Experiments are performed on a computer with Intel(R) Xeon(R) 3.0GHz CPU.

\subsection{Experiment Result}

Figure~\ref{fig1} shows that our parallel algorithm can gain near linear speed up. With 10 threads, our framework brings 4-fold to 6-fold faster speed than that with only 1 thread. Table~\ref{table2} shows the speed up in our benchmark datasets with 1,4 and 10 threads.

Figure~\ref{fig2} also shows that our parallel framework has no loss in accuracy/F-score or convergence rate compared with single-thread baseline. Table~\ref{table1} indicates that our framework does not hurt large margin algorithm because the difference of results is very small. In other words, there is barely interference among threads, and the strong robustness of large margin algorithm ensures no loss in performance under the parallel framework.

\section{Conclusions}

We propose a generic online parallel learning framework for large margin models. Our experiment concludes that the proposed framework has no loss in performance compared with baseline while the training speed up is near linear with increasing running threads.

\section{Acknowledgements}

This work was supported in part by National Natural Science Foundation of China (No. 61673028), and National High Technology Research and Development Program of China (863 Program, No. 2015AA015404). Xu Sun is the corresponding author of this paper.

\bibliography{acl2016}
\bibliographystyle{acl_natbib}

\end{document}